\documentclass{bmvc2k}

\usepackage{times}
\usepackage{epsfig}
\usepackage{graphicx}
\usepackage{amsmath}
\usepackage{amssymb}
\usepackage{algorithm}
\usepackage[noend]{algpseudocode}
\usepackage{varwidth}
\usepackage{mwe}
\usepackage{newunicodechar}

\title{Improved Image Segmentation via Cost Minimization of Multiple Hypotheses}

\addauthor{Marc Bosch}{marc.bosch.ruiz@jhuapl.edu}{1}
\addauthor{Christopher M. Gifford}{Christopher.Gifford@jhuapl.edu}{1}
\addauthor{Austin G. Dress}{Austin.Dress@jhuapl.edu}{1}
\addauthor{Clare W. Lau}{Clare.Lau@jhuapl.edu}{1}
\addauthor{Jeffrey G. Skibo}{Jeffrey.Skibo@jhuapl.edu}{1}
\addauthor{Gordon A. Christie}{Gordon.Christie@jhuapl.edu}{1}

\addinstitution{
 Applied Physics Laboratory\\
 The Johns Hopkins University\\
 Laurel, MD, USA
}

\runninghead{Bosch et al.}{Improved Image Segmentation via CMMH}

\begin{document}

\maketitle

\begin{abstract}
Image segmentation is an important component of many image understanding systems. It aims to group pixels in a spatially and perceptually coherent manner. Typically, these algorithms have a collection of parameters that control the degree of over-segmentation produced. It still remains a challenge to properly select such parameters for human-like perceptual grouping. In this work, we exploit the diversity of segments produced by different choices of parameters. We scan the segmentation parameter space and generate a collection of image segmentation hypotheses (from highly over-segmented to under-segmented). These are fed into a cost minimization framework that produces the final segmentation by selecting segments that: (1) better describe the natural contours of the image, and (2) are more stable and persistent among all the segmentation hypotheses. We compare our algorithm's performance with state-of-the-art algorithms, showing that we can achieve improved results. We also show that our framework is robust to the choice of segmentation kernel that produces the initial set of hypotheses.
\end{abstract}

\section{Introduction}
\label{sec:intro}
\begin{figure}[ht]
   \bmvaHangBox{\fbox{\includegraphics[scale=0.52]{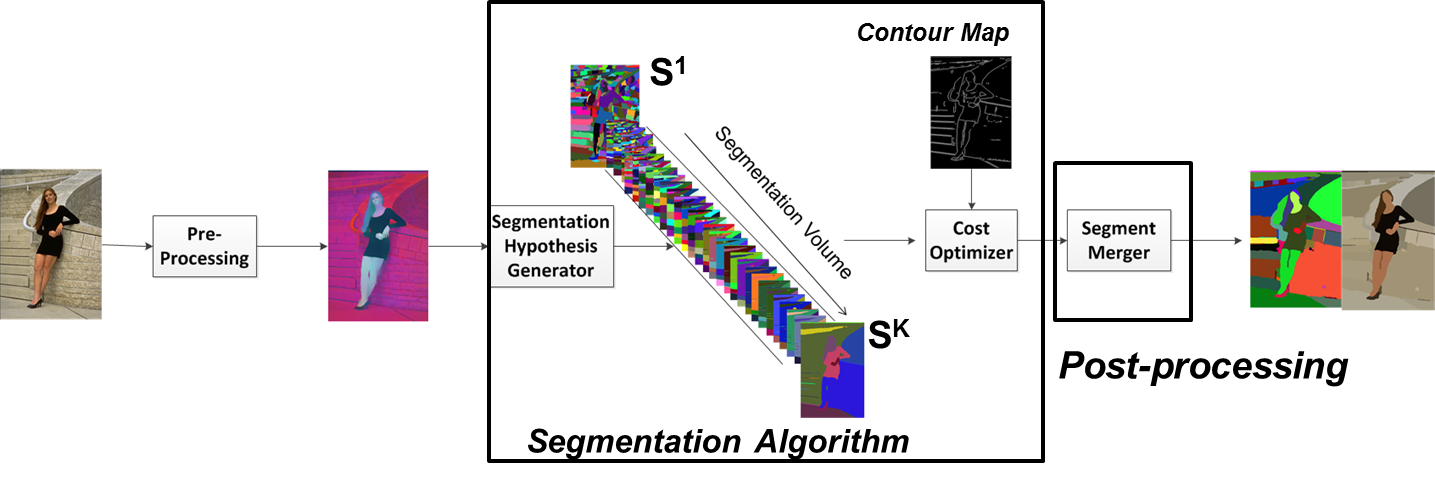}}}
   \caption{System block diagram of our proposed algorithm. It consists of three stages: (1) Pre-processing (color space and de-noiser), (2) Segmentation Block (segmentation hypothesis generator and cost optimizer), and (3) segmentation merger post-processing.}
\label{fig:overview}
\end{figure}

Image segmentation is a fundamental problem in computer vision and image processing. Its goal is to coherently group pixels in the image according to one or several visual attributes creating a mapping between objects/parts and pixels. Broadly speaking, it involves integrating (spatial) neighboring features using some similarity measure according to some fitting, partitioning, or merging criteria. In many cases, segmentation is seen as one of the initial steps (\textit{e.g.} region proposal step) of more high-level task algorithms such as object localization, identification, and tracking.\\
Over the years many segmentation algorithms have been proposed. Some focus on pixel grouping strategies, including graph partitioning/merging schemes \cite{D-bib:fh04,D-bib:ncut}, or iterative clustering like Mean-Shift \cite{D-bib:ms}. Other methods rely on the detection of natural contours and edges that ignore smooth variations in the image, focusing on the areas where visual features undergo rapid change \cite{D-bib:ac, D-bib:acchan, D-bib:cov, D-bib:ren, D-bib:bel, D-bib:ccp}. These are regarded as core segmentation algorithm kernels, or segmenters. More recently, algorithms that work at different hierarchies or scales have been proposed. All of these algorithms create a base layer with a segmenter that produces superpixels. These are grouped together by having an algorithm that measures their degree of similarity using some layered affinity representation~\cite{D-bib:selectivesearch, Chen2016,D-bib:mlss,D-bib:sas, D-bib:donoser}.\\
Segmentation algorithms typically have a collection of parameters, $\vec{a}$, that control their sensitivity to noise, illumination variations, pose changes, background contrast, and/or object class variation. Examples include number of clusters, spatial and color similarity thresholds, number of iterations, bandwidth, and many more \cite{D-bib:fh04,D-bib:ncut,D-bib:ac,D-bib:ms}. Different combinations and choices of such parameters, $\vec{a}$, tend to be more or less effective when applied to different scenes (over-segmentation vs. under-segmentation). It would require an oracle agent to set the right parameters beforehand to obtain the closest segmentation to human perception.\\
We observed that, in general, there is a combination of such parameters that produce a good result on a per image basis. Therefore, if we were to compute multiple segmentation hypotheses by perturbing the parameters, then it is likely that a good grouping could be found in the scene. This may require using segments from different segmentation hypotheses for a given image.\\
In this work, we propose an image segmentation framework that uses low-level cues (\textit{i.e.} color and intensity) and works on the entire space of segmentation hypotheses or ``segmentation volume'' (\textit{i.e.} a set of segmentations generated from different parameter choices given a segmentation kernel). See Figure \ref{fig:overview}. The algorithm looks, within the pool of hypotheses, for the most stable grouping that best describes the natural contours present in the scene. We work with multiple segmentations, as we acknowledge that one set of parameters cannot work consistently for different scenes and images. Our approach defines a cost function according to two criteria: (1) segments that change constantly and abruptly in the segmentation volume receive larger penalties, and (2) segments that do not match with natural image contours should be discouraged. This function is minimized in order to fuse the hypotheses and obtain a final segmentation. As shown in Figure \ref{fig:overview}, our segmentation hypothesis generator and cost optimizer blocks are complemented with simple pre-processing and post-processing operations. Our code is available \footnote{https://github.com/pubgeo/cmmh\_segmentation}.\\
The remainder of the paper is organized as follows: In section~\ref{sec:isa}, we describe our approach including the core algorithm as well as pre-processing and post-processing steps. In section~\ref{sec:eval}, we show our experimental results, and how our method compares to other state-of-the-art solutions. Finally, we conclude the paper by offering some remarks from our experimental observations.
\section{Image Segmentation Algorithm}
\label{sec:isa}
This section describes the proposed segmentation approach. We start by outlining a few concepts to lay the groundwork, and then detailing the different algorithm components.\\
We define a segmentation $\mathbf{S}={S_1,S_2,...,S_m}$ with $m$ segments as follows: Given an image $I$, $\mathbf{S}$ is the particular grouping of pixels obtained from processing a similarity measure by an algorithm using parameters $\vec{a_i}$, \textit{i.e.} passing the image through a segmenter with a set configuration. We denote this as $\mathbf{S(}\vec{a_i}\mathbf{)}$.\\
This formulation allows us to represent multiple segmentation hypotheses as a function of these parameters, $\mathbf{S(}\vec{a_1}\mathbf{)},\mathbf{S(}\vec{a_2}\mathbf{)},...,\mathbf{S(}\vec{a_K}\mathbf{)}$. For simplicity, hereon, we refer to $\mathbf{S(}\vec{a_i}\mathbf{)}$ as $\mathbf{S^i}$. By varying such parameters, the segmented regions vary in size (number of pixels) and/or in number. Therefore, $\mathbf{S^1}={S_1^1,...,S_{m_1}^1}; \mathbf{S^2}={S_1^2,...,S_{m_2}^2};...;\mathbf{S^K}={S_1^K,...,S_{m_K}^K}$, where $m_1,m_2,...,m_K$ represent the number of segments, and $S_j^1,S_j^2,...,S_j^K$ map into different pixels.\\
Note that our concept of multiple segmentation hypotheses differs from the notion of hierarchy of segmentations. A hierarchy of segmentations starts with a fine set of superpixels that are iteratively merged to create coarser groupings. 
On the contrary, our set of segmentation hypotheses are independently created by a segmenter, or segmentation kernel, by modifying the parameters $\vec{a_i}$, that control the pixel affinity sensitivity. As hierarchical approaches, our approach also tends to produce finer and coarser partitions, but coarse partitions are not necessarily obtained from merging finer partitions.
\subsection{Algorithm Overview}
With these definitions in mind, our algorithm starts by proposing a pool of possible segments in the parameter space ($\vec{a_1},...,\vec{a_i},...,\vec{a_K}$), \textit{i.e.} $\mathbf{S^1},...,\mathbf{S^i},...,\mathbf{S^K}$. In particular, our choice of segmenter is a graph-based segmentation based on the minimum spanning forest model introduced by Felzenszwalb and Huttenlocher in \cite{D-bib:fh04}, hereon referred to as \textit{FH}. Each set $\mathbf{S^i}$ is obtained by modifying the threshold, $\kappa$, that controls the degree to which two minimum spanning tree components of the graph differ from each other. These segmentation hypotheses form a 3-D space, \textit{i.e.} a segmentation volume, where one dimension is represented by $\kappa_i$ and the other two dimensions are generated by the coordinates of the image lattice (as shown in figure \ref{fig:overview}). Thus, each pixel belongs to many segments (one per each segmentation $\mathbf{S^i}$). The decision of which is the best segment for a given pixel follows next.\\ 
A cost minimization framework is proposed to integrate all the hypotheses and create a final segmentation. A pixel is assigned a penalty given the set of segments it belongs to for a collection of segmentations $\mathbf{S^1},...,\mathbf{S^K}$. The resulting final segmentation assigns the segment with the minimum penalty (while enforcing spatial uniformity) to each pixel.
\subsection{Cost Calculation and Minimization}
\begin{figure}[ht]
   \bmvaHangBox{\fbox{\includegraphics[scale=0.28]{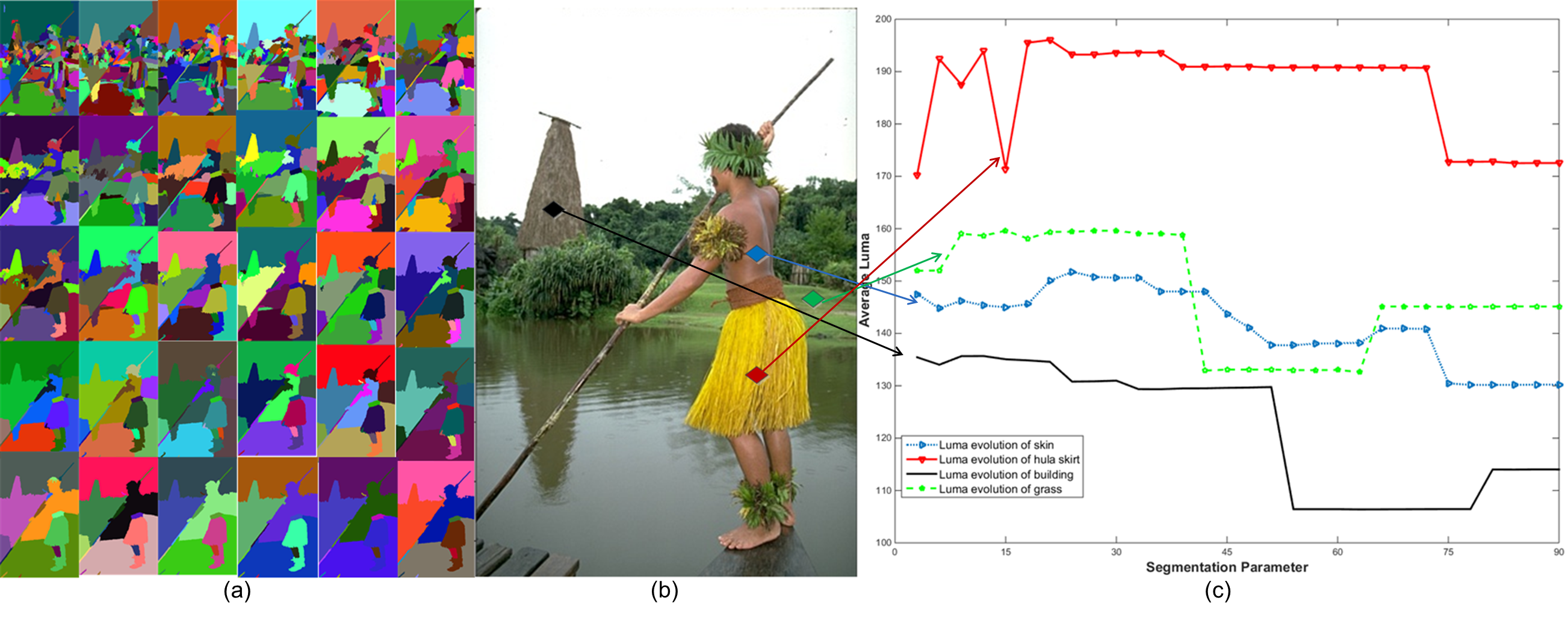}}}
   \caption{Example of the evolution of average luma value in selected pixels as the algorithm covers the segmentation parameter space.}
\label{fig:1dsignal}
\end{figure}

Natural images are spatially coherent, \textit{i.e.} neighboring pixels of the same object tend to have similar visual attributes. This observation allows us to impose the following criteria when designing the algorithm: a gradual variation in the parameter $\vec{a_i}$ space should in turn create a gradual change in the pixel coverage of each segment. By sorting the segmentations $\mathbf{S^1},...,\mathbf{S^K}$ in such a way that consecutive segmentations are close in parameter space ($\vec{a_1},...,\vec{a_K}$), consecutive semantically similar segmentations will be produced. The parameter range can be set so that it covers as much of the solution space as possible. In other words, $\mathbf{S^1}$ represents a highly over-segmented image and $\mathbf{S^K}$ an under-segmented image. This enforces that, under the assumption that similar pixels tend to belong to the same object, small variations of $\vec{a_i}$ will produce similar segments.\\
In order to generate the final segmentation, we have designed a cost minimization framework where each pixel is assigned a penalty according to the following two conditions: (1) segment boundaries should align with strong natural edges, and (2) small changes in parameter space should not easily transition from over-segmentation to under-segmentation situations, giving a sense of segment stability for small parameter change. Thus, our cost function has the general form: 
\begin{equation}
\label{eq:cost}
cost(\vec{a_i})=\omega_1 \cdot Cost_1(\vec{a_i})+\omega_2 \cdot Cost_2(\vec{a_i})+Reg
\end{equation}
where the first condition about natural edges is modeled by the $Cost_1$ term, and the second condition's cost will be represented by $Cost_2$. $\omega_1$ and $\omega_2$ parameters' role is to find a balance between the two. $Reg$ is the regularization term (smoothness prior) that will encourage segment coherency among neighboring pixels. Therefore, we want to select the segment configuration that minimizes $cost(\vec{a_i})$.\\
In order to model assumption 1, we propose to compare the edge map $E_S$ resulting from each segmentation $\mathbf{S^i}$ with a reference binary edge map $E_R$. We chose the structured edge detection algorithm~\cite{D-bib:dollar} to construct the reference binary edge map. We measure the degree of agreement between $E_S$ and $E_R$ for the $j_{th}$ segment, $S_j^i$, of segmentation $\mathbf{S^i}$ as:
\begin{equation}
Cost_e(r,c)=\frac{\sum_{(r,c \in S_j^i)}|E_S(r,c)-E_R(r,c)|}{\sum_{(r,c \in S_j^i)}E_R(r,c)}
\end{equation}
Note that we define $E_S$ as a binary image with value equal to one for pixels where at least one of its four immediate neighbors belongs to a different segment, and zero otherwise. $r,c$ are row and column indexes.\\
Our second condition is modeled by measuring, for each pixel, the differences between the average color of the segment it belongs to and its two immediate neighboring segments in the segmentation volume resulting from a different segmentation parameter configuration. We want the cost $Cost_c$ to indicate change in the pixel composition of each segment compared to their counterparts in the segmentation volume. Specifically, we set the cost of a pixel at location $(r,c)$ belonging to segment $S_j^i$ as:
\begin{align}
\begin{split}
Cost_c(r,c,\vec{a_i}) =2\mu_{I1}^{\vec{a}_i}(r,c)-\mu_{I1}^{\vec{a}_{i-1}}(r,c)-\mu_{I1}^{\vec{a}_{i+1}}(r,c) + \\
	+2\mu_{I2}^{\vec{a_i}}(r,c)-\mu_{I2}^{\vec{a}_{i-1}}(r,c)-\mu_{I2}^{\vec{a}_{i+1}}(r,c) + \\
	+2\mu_{I3}^{\vec{a_i}}(r,c)-\mu_{I3}^{\vec{a}_{i-1}}(r,c)-\mu_{I3}^{\vec{a}_{i+1}}(r,c)
\end{split}
\end{align}
with parameter $\mu$ denoting the average value of $I1,I2,I3$, which are the three color space planes, at pixel location $(r,c)$. $i$ is the index in the segmentation volume. The average value is calculated using all pixels of segment $S_j^i$. Figure \ref{fig:1dsignal} shows several examples of the evolution of the $\mu$ parameter as it traverses the segmentation volume. Figure \ref{fig:1dsignal}.c shows four locations in the image. Large penalties should be given for segments that show a noticeable change. Such changes indicate that the current segment has changed in composition with respect to the next segment for the same location. Small penalties should be given to those segments that show stability with respect to its immediate consecutive neighbors indicating segment consistency.\\
Both $Cost_c$ and $Cost_e$ are subtracted from their maximum values to obtain the penalty measures: $\Psi_c$, and $\Psi_e$, \textit{i.e.} $\Psi_c=Max(Cost_c) - Cost_c$ and $\Psi_e=Max(Cost_e) - Cost_e$. And finally, they are normalized to each be within range [0,1].\\
The combination of both cost measures takes into account segment consistency, and degree of agreement with strong natural edges. Following our formulation in Equation \ref{eq:cost} and replacing $Cost_1$ by $\Psi_e$ and $Cost_2$ by $\Psi_c$, our two-criteria cost function can be expressed as:
\begin{equation}
\label{eq:cost}
cost(\vec{a}_i)=\omega_c\cdot(\Psi_c(\vec{a}_i)) + \omega_e\cdot (\Psi_e(\vec{a}_i)) + \lambda\cdot\sum_{r_n,c_n\in N(r,c)}|\alpha(r,c)-\alpha(r_n,c_n)|
\end{equation}
where $\omega_c$ and $\omega_e$ are weights to indicate the importance of each cost measure, $\alpha$ is the index of the segmentation hypothesis ($\alpha=i)$, and $N(r,c)$ are the neighboring pixels. The regularization term determines how hypothesis support is aggregated. Finally, spatial smoothness to further encourage neighboring pixels being assigned to the same segment is done by applying a median filter to the cost function component $\omega_c\cdot\Psi_c(\vec{a}_i) + \omega_e\cdot\Psi_e(\vec{a}_i)$. A $5\times 5$ median filter is used.\\
Proper $\omega_c$ and $\omega_e$ have to be estimated to account for the degree of confidence each cost has in the pool of segmentation hypotheses. We repurpose the model proposed by Khelifi \textit{et al.}~\cite{D-bib:khelifi} for our problem. The degree of variation or uncertainty of each cost term across all segmentations is used as a measure of criteria importance. Such degree of uncertainty is measured by estimating the entropy of each cost term according to:
\begin{align}
\begin{split}
H_c &=-(1/log(n_c))\cdot\sum_{r=1...n_c} P(\Psi_c(r))log(P(\Psi_c(r)))\\
H_e &=-(1/log(n_e))\cdot\sum_{r=1...n_e} P(\Psi_e(r))log(P(\Psi_e(r)))
\end{split}
\end{align}
We assume uniform probability distribution when computing the entropy. This is $P(\Psi_c(r))=\frac{\sum n_c^r}{n}$ and $P(\Psi_e(r))=\frac{\sum n_e^r}{n}$ respectively, with $n_c^r$ and $n_e^r$ being the number of pixels with cost value equal to $\Psi_c(r)$ and $\Psi_e(r)$.\\
Finally, $\omega_c$ and $\omega_e$ in equation~\ref{eq:cost} are calculated as:
\begin{align}
\begin{split}
\omega_e &=\frac{1-H_c}{(1-H_e)+(1-H_c)}\\
\omega_c &=\frac{1-H_e}{(1-H_e)+(1-H_c)}
\end{split}
\end{align}
We minimize the cost function $cost(\vec{a_i})$, eq.~\ref{eq:cost}, by applying the Loopy Belief Propagation (LBP) algorithm. The optimization is over the segmentation hypothesis index $i$. It selects the segment with the smallest cost for a given pixel given all the segmentation hypothesis. We use the min-sum algorithm for LBP. Therefore, given pixel at position $(r,c)$ passing a message to his neighboring pixel $(r',c')$ and hypothesis index $i$, the min-sum belief message between these two pixels is calculated according to:
\begin{align}
\begin{split}
msg(r,c,r',c',i)=min_j(\omega_c\cdot(\Psi_c(\vec{a}_j)) + \omega_e\cdot (\Psi_e(\vec{a}_j))+\lambda\cdot|i-j|+\\
+\sum_{r_n,c_n\in N(r,c) \setminus(r',c')}msg(r_n,c_n,r,c,j))
\end{split}
\end{align}
The final belief for pixel $(r',c')$ using its 4-immediate neighbors, following formulation in eq.~\ref{eq:cost}, consists of:
\begin{equation}
belief(r',c',i)=cost(\vec{a}_i)+\sum_{r_n',c_n'\in N(r',c')}msg(r_n',c_n',r',c',i)
\end{equation}
LBP, then attempts to find the segmentation hypothesis index that minimizes the $belief$ for each pixel.

\subsection{Pre and Post-Processing}
\label{ssec:iprep}
Our image pre-processing step consists of a color space transformation from \textit{RGB} to \textit{Lab} space, and an image denoising step applied to each channel separately. The denoiser used is a weighted least squares image decomposition framework introduced in \cite{D-bib:fgs}.\\
Our post-processing is a simple segment merger algorithm based on segment size. After each pixel has been assigned to a segment, small segments are merged to adjacent ones until the combined area exceeds a threshold, $Th_1$.
\section{Evaluation}
\label{sec:eval}
\subsection{Experimental Results}
We have conducted our experiments on the BSDS 300 dataset \cite{D-bib:bsds300}. This dataset consists of 300 images and five sets of human-labeled segmentations. Results are reported after averaging the performance metrics among all five available human segmentation annotations. These performance metrics attempt to describe how close or similar a segmentation result is to the human versions (ground truth). We used four metrics, namely: (1) Probabilistic Rand Index (PRI), which computes the percentage of pixel-pair labels correctly assigned \cite{D-bib:pri}. The larger the PRI value, the closer the segmentation to the human ground truth. (2) Boundary Displacement Error (BDE), which measures the average pixel location error between segment boundaries \cite{D-bib:bde}. In this case, a lower BDE value indicates better performance. (3) Variation of Information (VOI), which attempts to measure the extent to which one segmentation explains the other. Lower VOI values indicate greater similarity with the ground truth \cite{D-bib:voi}. Finally, (4) Segmentation Covering (COV) evaluates the degree of coverage (segment-wise) of the segmentation algorithm with respect to the ground truth \cite{D-bib:cov}. Larger COV values indicate better overall performance.
\subsubsection{Performance Comparison}
Table \ref{tab:results} summarizes the main results of our experimentation. Algorithms are listed in terms of BDE performance. We compared our algorithm with popular algorithms for image segmentation. The first algorithm is the method presented in \cite{D-bib:cov}, \textit{gPb-owt-ucm}. We used the implementation made available by the authors \footnote{https://www2.eecs.berkeley.edu/Research/Projects/CS/vision/grouping/resources.html}. The second algorithm compared is contour-guided color palettes, \textit{CCP}  \cite{D-bib:ccp}, with the author's implementation available\footnote{https://github.com/fuxiang87/MCL\_CCP}. \textit{CCP} samples the image data around boundaries, and uses this information to guide a Mean-Shift algorithm. The third algorithm is a very popular region proposal algorithm, \textit{Selective Search (SS)}\footnote{http://homepages.inf.ed.ac.uk/juijling/\#page=software}, that uses \textit{FH} segmentation to generate an initial pool of segment hypotheses that are hierarchically integrated to each other with higher dimensional descriptions. There is a similarity measure that controls the superpixel grouping process. In order to calculate segmentation metrics on \textit{SS} segmentations, we stop the hierarchical grouping step at different levels of similarity given by fixed thresholds. In this paper, we list the \textit{SS} layer with best performing grouping in the hierarchy based on best BDE metric. Finally, \textit{FH} and Mean-Shift algorithms are also compared.\\
As mentioned earlier for the case of \textit{SS}, some of these algorithms \cite{D-bib:cov} involve choice of scales or other parameters. In order to establish a fair comparison, the results reported correspond to the segmentation results that produced the best results in terms of average BDE for all algorithms for all 300 images of the dataset. Also note that, as mentioned earlier, we have added pre-processing and post-processing (section \ref{ssec:iprep}), to our algorithm. These steps have also been applied to each of the benchmarking algorithms. In other words, the input image has been denoised and converted into \textit{Lab} color space, and small segment have been merged to adjacent ones. We have also included visual comparisons (see Figure~\ref{fig:comparisons} for details).
\subsubsection{Segmentation Kernel Comparison}
\begin{figure}
  \begin{center}
\begin{tabular}{cc}
\bmvaHangBox{\fbox{\includegraphics[width=5.25cm]{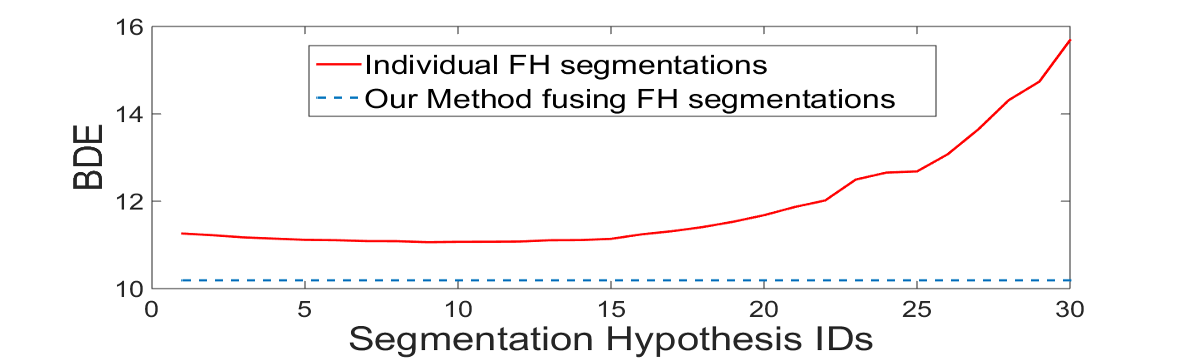}}}{(a)}& 
\bmvaHangBox{\fbox{\includegraphics[width=5.25cm]{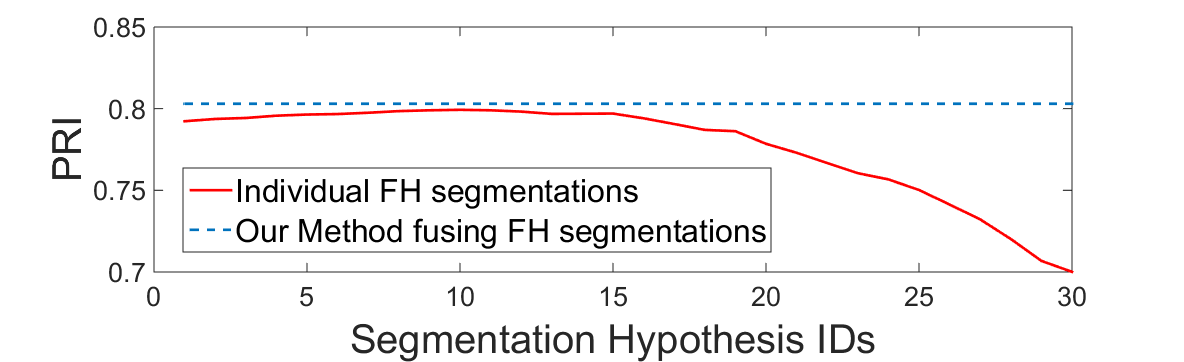}}}{(b)}\\ 
\bmvaHangBox{\fbox{\includegraphics[width=5.25cm]{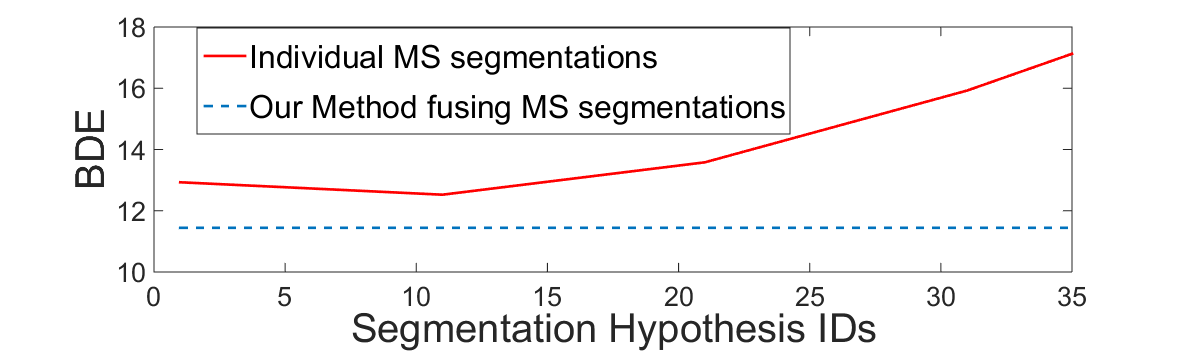}}}{(c)}& 
\bmvaHangBox{\fbox{\includegraphics[width=5.25cm]{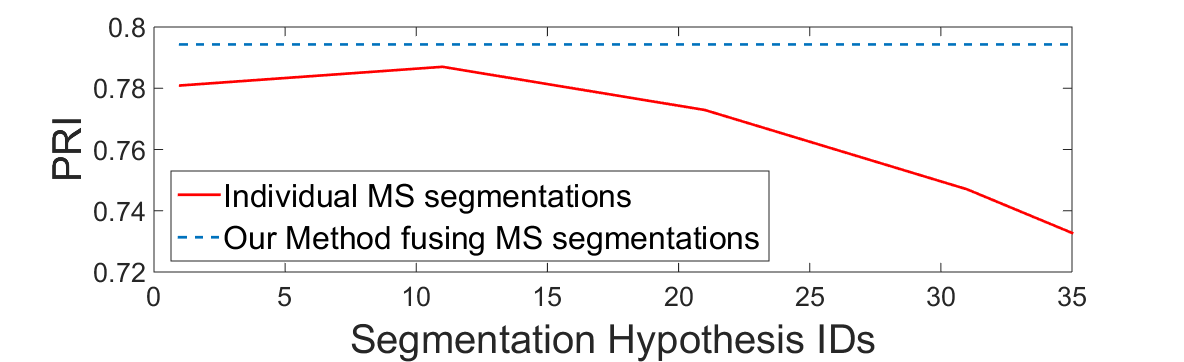}}}{(d)}\\ 
\end{tabular}
\caption{BDE and PRI performance curves using the BSDS300 dataset. In red, performance of all segmentation hypotheses. In blue (dotted), the resulting final segmentation with our approach. Experiments conducted for two different segmentation kernels: FH~\cite{D-bib:fh04} (a) and (b); MS~\cite{D-bib:ms} (c) and (d).}
\label{fig:bdepri}
  \end{center}
 \end{figure}

Equally relevant was to investigate how our approach improves from the individual segmentation hypotheses available before cost optimization. Figure \ref{fig:bdepri}.a and Figure \ref{fig:bdepri}.b show the results of each segmentation hypothesis available at the cost optimization stage, and the final segmentation in terms of BDE and PRI. We can see that for both measures the final segmentation achieves better results that any of its individual parts. The convex behavior of the individual plot indicates that the parameter space has been fully exploited in order to cover many segment possibilities (under- and over-segmentations). For completeness, we wanted to measure the extent to which our framework is robust to other segmentation kernels, and not tied to the specific choice of \textit{FH}. For this purpose we have also evaluated our algorithm with a different segmentation kernel. We replaced the \textit{FH} algorithm with another popular segmentation algorithm: the Mean-Shift, \textit{MS}, clustering algorithm. Figure \ref{fig:bdepri}.c and figure \ref{fig:bdepri}.d show the results of these experiments. We observe a similar trend as in the case of \textit{FH}, even though the overall performance of \textit{MS} for this dataset does not match as well with the human annotation. In both cases we observe that our approach can obtain closer segmentation results to ground truth than any of the individual hypotheses, and therefore outperforms manual parameter tuning.
\subsection{Implementation Details}
The following are the particular settings used in our experiments. These have been set once, at the beginning. In the optimization step, we set $\lambda$ equal to $1e^{-4}$. We used the following configuration for the image smoothing pre-processing step \cite{D-bib:fgs}: $sigma=0.01$, $lambda=900$, 4 iterations, and $attenuation=4$. In the post-processing step, we set $Th_1$ equal to 100 pixels as the minimum segment size (\ref{ssec:iprep}).  These parameters are common to all the other algorithms we benchmarked.
\begin{table}[ht]
\caption{Evaluation Results. Comparison of various segmentation approaches using the BSDS 300 Dataset. For BDE and VOI, lower values represent better performance. For PRI and COV, higher values represent closer to human annotation. Table has been sorted in terms of BDE performance.}
\centering
\begin{tabular}{c c c c c}
\hline\hline
Method & BDE & PRI & VOI & COV\\
\hline
\hline
Ours using FH & 10.20 & 0.80 & 2.16 & 0.56\\
\hline
CCP~\cite{D-bib:ccp} & 10.21 & 0.79 & 2.89 & 0.45\\
\hline
FH~\cite{D-bib:fh04} & 11.06 & 0.79 & 2.26 & 0.54\\
\hline
gPb-owt-ucm~\cite{D-bib:cov} & 11.32 & 0.79 & 2.68 & 0.49\\
\hline
Selective Search~\cite{D-bib:selectivesearch} & 12.01 & 0.77 & 2.72 & 0.48\\
\hline
MS~\cite{D-bib:ms} & 12.52 & 0.78 & 2.14 & 0.54\\
\hline
\end{tabular}
\label{tab:results}
\end{table}

\begin{figure}[ht]
   \bmvaHangBox{\fbox{\includegraphics[scale=0.48]{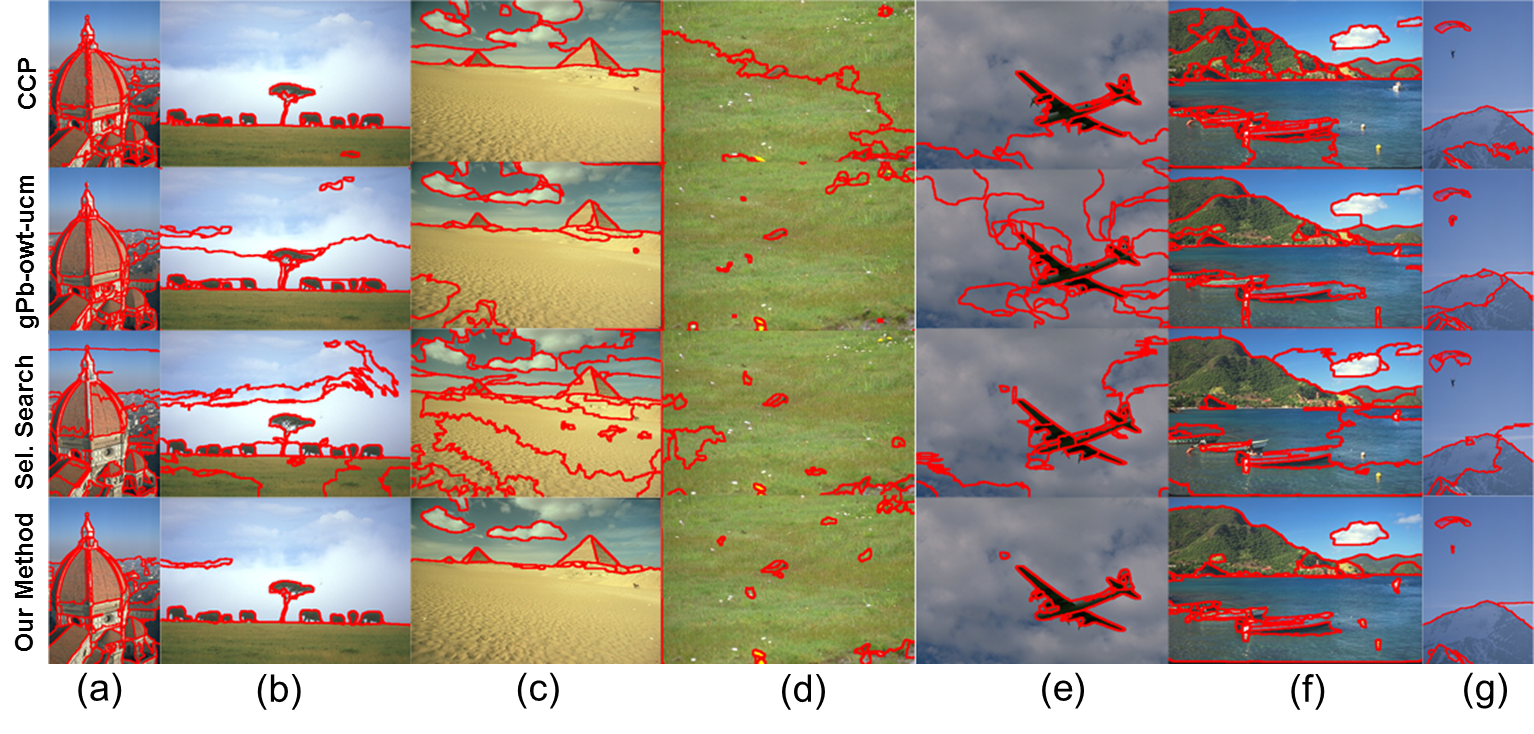}}}
   \caption{Visual examples of several segmentation methods using the BSDS 300 dataset. Note that same pre- and post-processing operations have been applied to all the algorithms.}
\label{fig:comparisons}
\end{figure}

\section{Results Discussion}
\label{sec:dis}
\begin{figure}[h]
\centering
   \bmvaHangBox{\fbox{\includegraphics[scale=0.63]{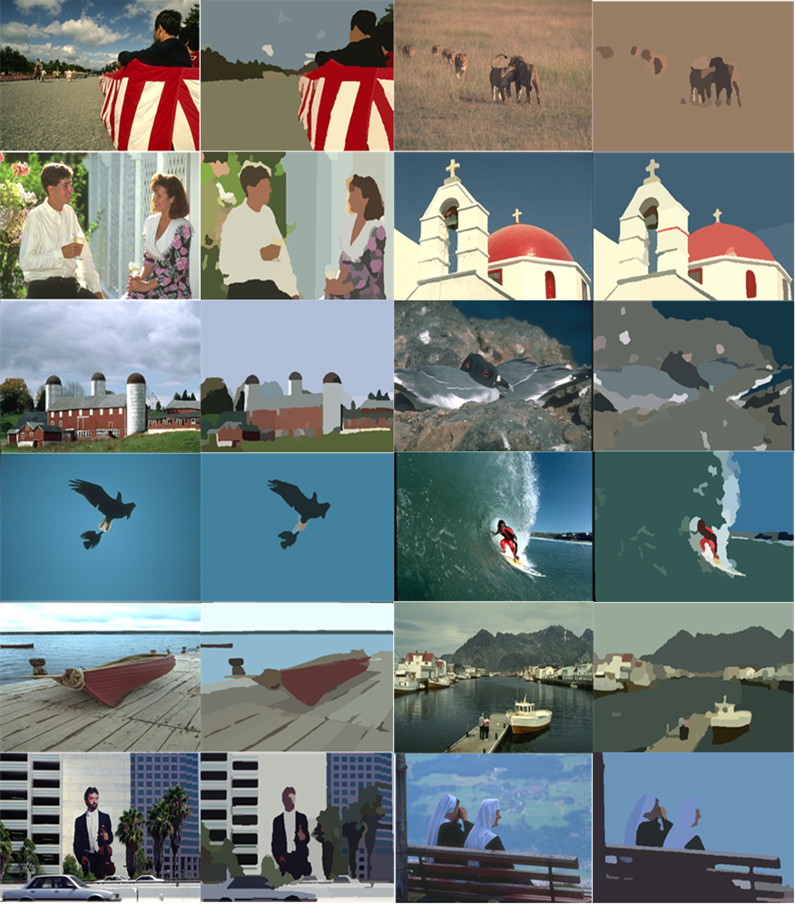}}}
   \caption{Segmentation results of the proposed method using the BSDS 300 dataset. Odd columns are the original input images, and even columns are our results. The average color value of the segment is used for visualizing results.}
\label{fig:visresquad}
\end{figure}

Table~\ref{tab:results} shows how the proposed method outperforms the other methods given our experimental setup for all metrics used. Our solution is capable of achieving low BDE numbers while keeping COV values high, indicating that it matches human-defined boundaries better while producing higher segmentation coverage. Figure~\ref{fig:comparisons} shows some of the benchmarked algorithms suffering from under-segmentation in certain situations (e.g. \textit{CCP} and \textit{gPb-owt-ucm} undersegment clouds in Figure~\ref{fig:comparisons}.c and Figure~\ref{fig:comparisons}.f respectively). \textit{SS}, for instance, suffers from both over- (Figure~\ref{fig:comparisons}.c the sand in the desert) and under-segmentation (Figure~\ref{fig:comparisons}.f mountain vs. water). Even though our approach also suffers from under- or over-segmentation, as can be seen in Figure~\ref{fig:comparisons}.e (plane), it does so at a lower extent. Our method better handles smaller objects without excessive over-segmentation, as can be seen in Figure~\ref{fig:comparisons}.b (elephants) and Figure~\ref{fig:comparisons}.g (paraglider). Again, all these results are obtained by the segmentations for each algorithm that produced the lower (better) BDE performance.\\
The results shown in Figure~\ref{fig:bdepri} seem to indicate that having an algorithm that can leverage segments from across the parameter space allows it to group pixels in a more coherent way than any of the individual parameter configurations alone. These results have been validated for more than one particular choice of the segmentation algorithm (\textit{i.e.} \textit{FH} and \textit{MS}), which makes the overall framework robust to the particular choice of the segmentation algorithm.\\
The design of our algorithm requires a collection of segmentation hypotheses to be computed. This translates into having to run a particular segmentation algorithm (\textit{e.g.} \textit{FH}) multiple times for different parameter choices. Our algorithm does not require these segmentations to be sequentially computed, as there is no feedback indication for the next choice in the parameter set. In this sense, it is a naive production of segments that are all independently computed. For this reason, parallelism can be exploited.\\
Finally, in Figure \ref{fig:visresquad} we have included a series of visual results to show the performance of our proposed algorithm on other BSDS 300 dataset images.
\section{Conclusion}
\label{sec:con}
In this work we have proposed a segmentation method that addresses the problem of a good choice of segmentation parameter(s). First, the method proposes a series of segmentation hypotheses using an off-the-shelf segmentation kernel creating a segmentation volume where each pixel is assigned to a set of segments. The second step estimates a cost function that measures how well segment boundaries match natural contours found in the scene, and how stable and persistent the segments are, given a collection of segmentation hypotheses. By minimizing such a cost function, we are able to obtain segmentations that are closer to human annotations than any of the individual hypotheses. We have also shown how our method is robust to the particular choice of segmenter. Finally, we showed that this method is capable of outperforming popular segmentation algorithms.



\bibliography{egbib}
\end{document}